\documentclass[10pt,twocolumn,letterpaper]{article}

\usepackage[pagenumbers]{cvpr} 

\usepackage[dvipsnames]{xcolor}

\definecolor{cvprblue}{rgb}{0.21,0.49,0.74}
\usepackage[pagebackref,breaklinks,colorlinks,citecolor=cvprblue]{hyperref}

\usepackage{amsmath,amsfonts,bm}

\def\eqref#1{equation~\ref{#1}}

\def\ceil#1{\lceil #1 \rceil}

\def\1{\bm{1}}

\def\vz{{\bm{z}}}

\DeclareMathAlphabet{\mathsfit}{\encodingdefault}{\sfdefault}{m}{sl}
\SetMathAlphabet{\mathsfit}{bold}{\encodingdefault}{\sfdefault}{bx}{n}

\newcommand{\softmax}{\mathrm{softmax}}

\usepackage{url}

\usepackage[utf8]{inputenc} 
\usepackage[T1]{fontenc}    
\usepackage{hyperref}       
\usepackage{url}            
\usepackage{booktabs}       
\usepackage{amsfonts}       
\usepackage{nicefrac}       
\usepackage{microtype}      
\usepackage{xcolor}         
\usepackage[utf8]{inputenc} 
\usepackage[T1]{fontenc}    
\usepackage{hyperref}      
\usepackage{url}           
\usepackage{booktabs}       
\usepackage{amsmath}       
\usepackage{nicefrac}     
\usepackage{microtype}      
\usepackage{multirow}      
\usepackage{graphicx}
\usepackage{graphics}
\usepackage{tabularx}
\usepackage{makecell}
\usepackage{caption}
\usepackage{wrapfig}
\usepackage{enumitem}
\usepackage{subcaption}
\usepackage{amsfonts}      
\usepackage{nicefrac}  
\usepackage[abs]{overpic}

\usepackage{enumitem}
\usepackage{algorithmic}
\usepackage{algorithm}
\usepackage{verbatim}
\usepackage{listings}
\usepackage{wrapfig}
\usepackage{mathtools}
\definecolor{codeblue}{rgb}{0.25,0.5,0.25}
\definecolor{codekw}{rgb}{0.85, 0.18, 0.50}
\usepackage{subcaption}
\usepackage{tikz}
\usepackage{comment}

\usepackage{times}
\usepackage{epsfig}
\usepackage{graphicx}
\usepackage{amsmath, bm}
\usepackage{amssymb}
\usepackage{multirow}
\usepackage{microtype}
\usepackage[normalem]{ulem}
\useunder{\uline}{\ul}{}
\usepackage{color}
\usepackage{nicefrac}
\usepackage{comment}
\usepackage{enumitem}
\usepackage{adjustbox}
\usepackage{mathabx}
\usepackage{tabu}
\usepackage{booktabs}
\usepackage{diagbox}
\usepackage{cuted}
\usepackage{capt-of} 
\usepackage{multicol}

\usepackage{color, colortbl}
\definecolor{Gray}{gray}{0.93}
\newcommand{\tablestyle}[2]{\setlength{\tabcolsep}{#1}\renewcommand{\arraystretch}{#2}\centering\scriptsize}
\usepackage[capitalize]{cleveref}
\crefname{section}{Sec.}{Secs.}
\Crefname{section}{Section}{Sections}
\Crefname{table}{Table}{Tables}
\crefname{table}{Tab.}{Tabs.}

\usepackage{color, colortbl}
\definecolor{Gray}{gray}{0.93}
\definecolor{orange}{rgb}{0.9,0.5,0}

\usepackage{hyperref} 
 \hypersetup{ 
     colorlinks=true, 
     linkcolor=red, 
     }

\usepackage[margin=4pt,font=small,labelfont=bf,labelsep=endash,tableposition=bottom]{caption}
\newcolumntype{H}{>{\setbox0=\hbox\bgroup}c<{\egroup}@{}}

\definecolor{mycolor}{HTML}{5a860e}

\def\paperID{483} 
\def\confName{CVPR}
\def\confYear{2024}
\begin{document}
\title{\Large ViR: Towards Efficient Vision Retention Backbones}
\author{Ali Hatamizadeh$^{1,*}$, Mike Ranzinger$^{1,*}$, Shiyi Lan$^{1}$, Jose M. Alvarez$^{1}$, Sanja Fidler$^{1,2}$, Jan Kautz$^{1}$\\
$^{1}$NVIDIA, $^{2}$University of Toronto\\
{\tt\small {\{ahatamizadeh,mranzinger\}@nvidia.com}
}}

\maketitle

\def\thefootnote{*}\footnotetext{Equal contribution.}

\renewcommand{\thefootnote}{\arabic{footnote}}

\begin{abstract}
Vision Transformers (ViTs) have attracted a lot of popularity in recent years, due to their exceptional capabilities in modeling long-range spatial dependencies and scalability for large scale training. Although the training parallelism of self-attention mechanism plays an important role in retaining great performance, its quadratic complexity baffles the application of ViTs in many scenarios which demand fast inference. This effect is even more pronounced in applications in which autoregressive modeling of input features is required. In Natural Language Processing (NLP), a new stream of efforts has proposed parallelizable models with recurrent formulation that allows for efficient inference in generative applications. Inspired by this trend, we propose a new class of computer vision models, dubbed Vision Retention Networks (ViR), with dual parallel and recurrent formulations, which strike an optimal balance between fast inference and parallel training with competitive performance. In particular, ViR scales favorably for image throughput and memory consumption in tasks that require higher-resolution images due to its flexible formulation in processing large sequence lengths. The ViR is the first attempt to realize dual parallel and recurrent equivalency in a general vision backbone for recognition tasks. We have validated the effectiveness of ViR through extensive experiments with different dataset sizes and various image resolutions and achieved competitive performance. 
Code: \url{https://github.com/NVlabs/ViR}.  
\end{abstract}  


    
\section{Introduction}
\begin{figure}[]
\centering

\resizebox{1.\linewidth}{!}{
\begingroup
\renewcommand*{\arraystretch}{0.3}
\begin{tabular}{c}

  \includegraphics[width=1\linewidth]{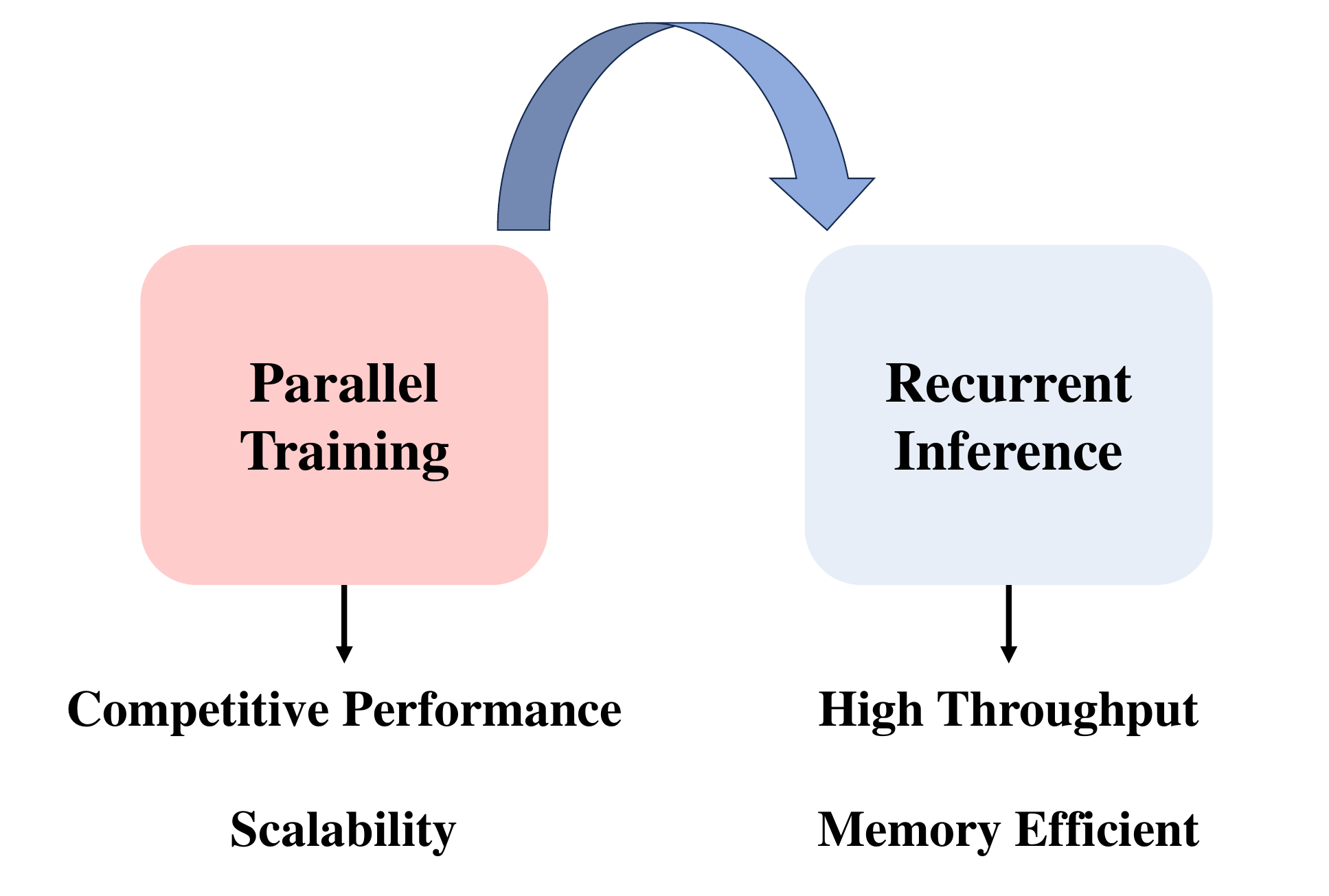}\vspace{3mm} \\ 
  (a) Overview of ViR framework.\vspace{3mm} \\
   \includegraphics[width=1\linewidth]{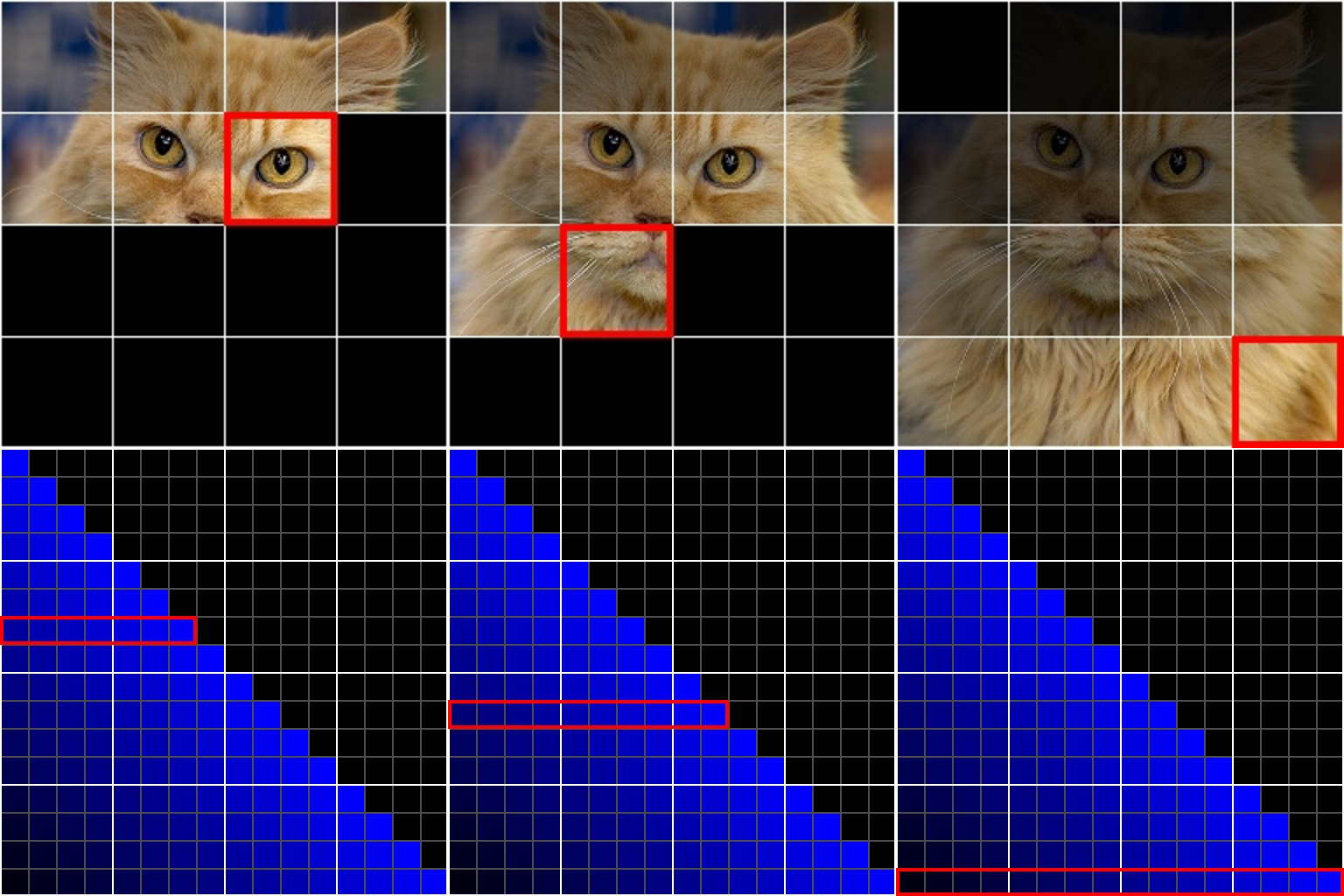}\vspace{3mm} \\ 
  (b) Effective receptive field and retention masks in ViR. \\
\end{tabular}
\endgroup
}
\caption{The proposed ViR enables dual parallel and recurrent formulations by using a retention mask. The ViR can be trained in the \textit{\textbf{parallel}} mode and achieve a competetive performance. The inference can leverage \textit{\textbf{recurrent}} or \textit{\textbf{chunkwise}} formulations to improve image throughput and memory efficiency. The effective receptive field and corresponding retention masks in ViR are visualized for a set of patches.}
\label{fig:overlaid_receptive}
\end{figure}
During the recent years, Transformers~\citep{vaswani2017attention} and their variants~\citep{devlin2018bert,dosovitskiy2020image} have shown competitive performance across multiple domains such as Natural Language Processing (NLP) and Computer vision. The main building block of Transformers is self-attention which allows for cross interaction among all input sequence tokens with each other. This scheme effectively captures short- and long-range spatial dependencies and imposes time and space quadratic complexity in terms of the input sequence length. The training parallelism of Transformers allows for competitive performance. However, the inference is slow and expensive due to the computational complexity. Recently, Retentive Network (RetNet)~\citep{sun2023retentive} and Receptance Weighted Key Value (RWKV)~\citep{peng2023rwkv} independently proposed novel model architectures that include the training parallelism of transformers and fast recurrent inference. The RWKV model uses linear channel-wise attention to relax the pairwise dot product bottleneck of vanilla self-attention. The RetNet, on the other hand, proposes the concept of retention with dual-form parallel and recurrent representations. It is noteworthy to mention that both RWKV and RetNet models are primarily proposed for autoregressive text generation. 

Although Convolutional Neural Networks (CNNs) have been commonly used as the de-facto architecture for various applications, the introduction of Vision Transformers~\citep{dosovitskiy2020image} (ViT) demonstrated the possibility of achieving State-of-the-Art (SOTA) performance with a similar model to the Transformers for NLP applications. As opposed to the autoregressive formulation in which tokens from left to right are processed at each step to predict the next value, ViT uses the entire token representations. In the case of long token sequences (\textit{e.g.} high-resolution images), processing the entire tokens may create a bottleneck due to the quadratic complexity of the self-attention layers. As a result, despite the competitive performance of ViT models, this limits their usage for applications that require real-time processing of high-resolution images (\textit{e.g.} autonomous vehicles).   

In this work, we explore the possibility of leveraging the duality of parallel and recurrent formulations to enable fast and memory-efficient deployment while maintaining the training parallelism with competitive performance. We introduce a new class of computer vision models, dubbed Vision Retention networks (ViR) which enables dual parallel and recurrent formulations. In addition, the combination of parallel and recurrent modes, referred to as chunk-wise formulation, enables an optimal combination of both modes based on specific run-time hyper-parameters (\textit{e.g.} batch size) and hardware requirements. Due to this formulation, the memory consumption in ViR model can then be decoupled from the sequence length, hence making it easier to process high-resolution images in an efficient manner. In order to improve the efficiency, we have redesigned the retention mechanism by removing the gated function. In addition, the proposed retention formulation is also generic and does not rely on any specific relative position embedding formulations. Our proposed ViR is the first attempt beyond generative applications for leveraging autoregressive vision-friendly retentive networks for recognition tasks (\textit{e.g.} image classification)

The summary of our contributions is as follows:

\begin{itemize}[noitemsep,nosep]
\item We introduce ViR, which is the first attempt at leveraging autoregressive retentive network with dual parallel and recurrent formulations for vision recognition tasks. We demonstrate that ViR can scale favorably to larger image resolutions in terms of image throughput and memory consumption.   

\item We propose 1D and 2D retention formulations, with desirable properties such as shift equivariance, that can be used for various downstream tasks (\textit{i.e.} detection and segmentation) with high-resolution images. 

\item We have validated the effectiveness of ViR by pretraining and finetuning on both ImageNet-21K and ImageNet-1K datasets for different models sizes to demonstrate the scalability of our proposed model as a general computer vision backbone.

\end{itemize}

\section{Related Work}

\paragraph{Vision Transformers}

ViT~\citep{dosovitskiy2020image} introduced a new paradigm to move away from the convolutional inductive biases towards a simpler model with minimal priors. The effectiveness of self-attention in modeling long-range spatial dependencies and the scalability of ViTs make them a great candidate as a backbone model for various vision tasks. However, the quadratic complexity of self-attention creates a bottleneck for fast deployment, especially for high-resolution images with longer sequence lengths. Swin Transformers~\citep{liu2021swin} proposed to compute self-attention in smaller partitioned windows to address this problem. Although this scheme improves the efficiency, the limited cross-region interactions across local windows may impact the performance. Independently, Pyramid Vision Transformer (PVT)~\citep{wang2021pyramid} introduced a hierarchical architecture, similar to Swin Transformer, that employs a patch embedding layer at the beginning of each stage and reduces the spatial dimension to improve the computational efficiency. On the other hand, Twins Transformer~\citep{chu2021twins} introduced a spatially separable self-attention mechanism that consisted of global sub-sampling and locally-grouped modules that can model both short and long-range interactions in an efficient manner. Several follow up efforts proposed to address this issue by introducing global~\citep{hatamizadeh2023global} or carrier~\citep{hatamizadeh2023fastervit} tokens and multi-axis grid attention~\citep{tu2022maxvit}.

\paragraph{Hybrid Models} 

In addition to these works, a stream of hybrid models (\textit{i.e.} CNN and ViT)~\citep{graham2021levit,Wu_2021_ICCV,yuan2021tokens} were proposed to improve the data efficiency and achieve competitive performance without considerably larger model sizes. Convolutional vision Transformer (CvT)~\citep{Wu_2021_ICCV} proposes the concept of convolutional token embedding layer which is integrated with a Transformer block in a hierarchical
architecture to improve the data efficiency and performance of the ViT models. In addition, Tokens-To-Token Vision Transformer (T2T-ViT)~\citep{yuan2021tokens} introduced a tailored transformation layer for aggregating nearby tokens which can be ten used as image priors for leveraging spatial correlations. Cross-covariance Image Transformer (XCiT)~\citep{ali2021xcit} proposed a transposed self-attention block for capturing the token interactions in feature channels space. In addition, by conditioning the position encoding on localized patch tokens, Conditional Position encoding Vision Transformer (CPVT)~\citep{chu2021conditional} achieved better performance on different recognition tasks such as image classification and object detection. Our proposed contributions in this work are orthogonal to these recent advances as ViR can benefit from a hybrid architecture as well as a window-based retention.

 \begin{figure*}[h]
    \centering
    \includegraphics[width=0.86\textwidth]{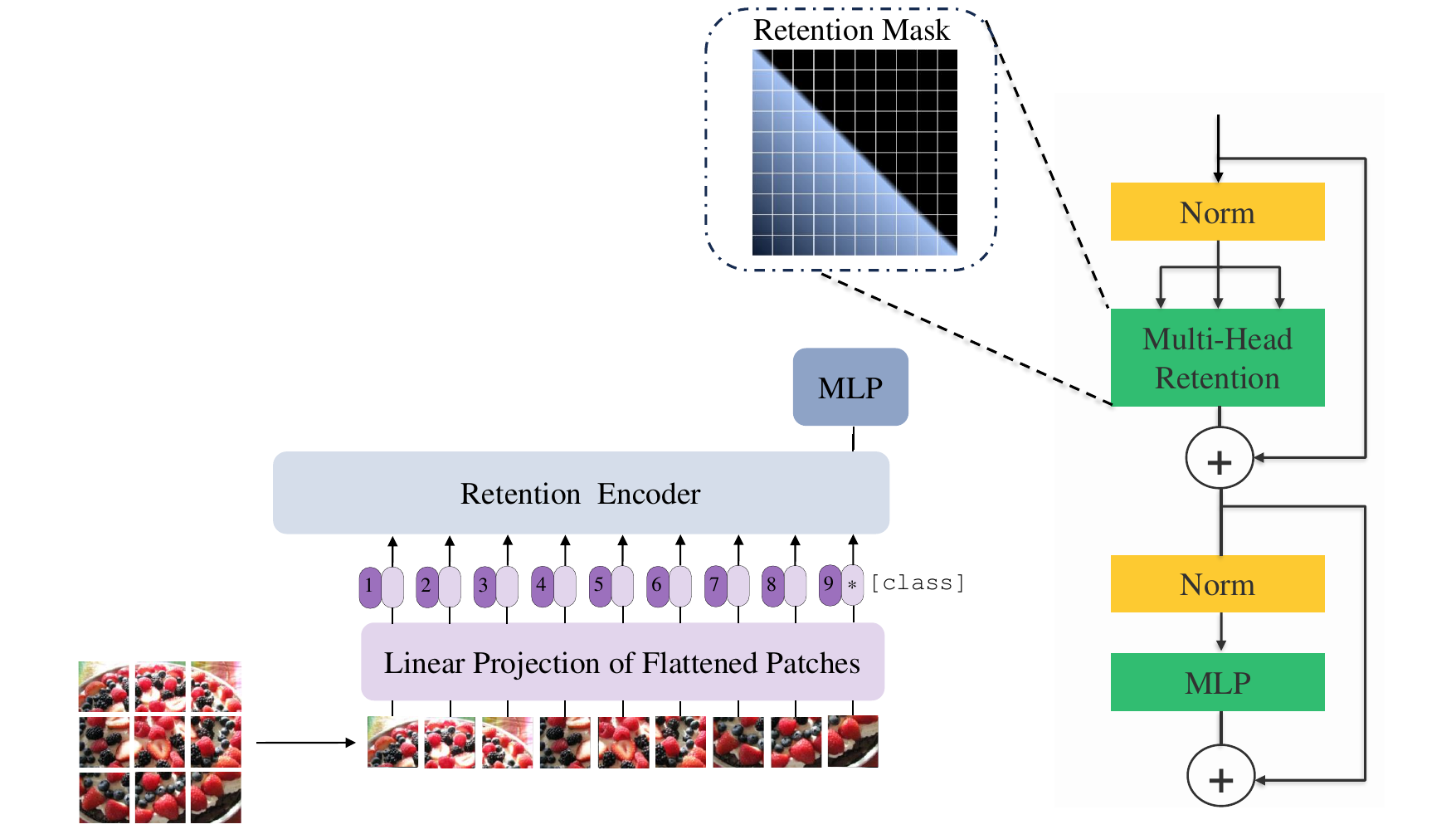}
    \caption{Overview of the architecture of ViR model. Similar to ViT, Flattened patches are linearly projected into a patch embedding. The position embedding are then added to the patch embedding and a class token is appended to this sequence. The retention encoder comprises of alternating Multi-Head Retention and MLP blocks. The MHR blocks use a causal decay mask. Please see the supplementary materials for detailed information regarding the architecture of H-ViR model.}
    \label{fig:vir_arch}
\end{figure*}

\paragraph{Autoregressive Models}
Deep Autoregressive models~\cite{greff2016lstm,van2016pixel,van2016conditional,chen2018pixelsnail,radford2018improving} have primarily been used for generative application and achieved great success in this domain. Most notably, PixelCNN~\citep{van2016conditional} and PixelRNN~\citep{van2016pixel} demonstrated that sequential pixel-by-pixel prediction can be effective in learning the explicit probability distribution for both discrete and continuous data while having better training stability compared to Generative Adversarial Networks (GANs)~\citep{goodfellow2014generative}. With the emergence of Transformers~\citep{vaswani2017attention}, several efforts ~\citep{parmar2018image, chen2020generative, cao2021image, chang2022maskgit} demonstrated the capability of autoregressive modeling at scale. However, the sequential nature of autoregressive decoding, which requires access to previously generated tokens, hinders the efficiency of such models. 

\paragraph{Self-attention Alternatives}
To address the quadratic computation complexity of self-attention, many efforts have proposed various approaches such as approximation of the $\softmax$ activation function~\citep{joulin2017efficient, gao2020design}, linear attention by using other kernels~\citep{wang2020linformer,katharopoulos_et_al_2020} to estimate the attention scores or computing the attention in the channel feature space~\citep{ali2021xcit}. However, the improved efficiency negatively impacts the performance of the model. Other efforts~\citep{zhai2021attention, gu2021efficiently} have also proposed to entirely replace the self-attention with other mechanisms. In particular, recently in NLP, RWKV~\citep{peng2023rwkv} and RetNet~\citep{sun2023retentive} proposed to redefine the Transformers to leverage the duality of parallel and recurrent formulation for training and inference. RWKV follows an attention-free formulation~\citep{zhai2021attention} but employs an exponential decay to enable the recurrent formulation. RetNet proposes to use multi-scale gated retention to maintain the expressivity of the contextual information and achieve competitive performance. Although our work is inspired by RetNet, it is aimed for computer vision, in particular recognition, and has a tailored retention mechanism and architecture redesign for optimal performance.

\section{Methodology}
\subsection{1D Retention}
In this section, we discuss the retention mechanism and its different formulations~\citep{sun2023retentive}. Consider an input sequence $\mathbf{X} \in \mathbb{R}^{|X|\times D}$ that will be encoded in an autoregressive manner. Given the query ($\mathbf{q_{n}}$), key ($\mathbf{k_{n}}$) and value ($\mathbf{v_{n}}$) in state $\mathbf{s_{n}}$, this sequence-to-sequence mapping can be written as
\begin{equation}
\begin{aligned}
\label{eq:rnn}
&\mathbf{s_n} = \gamma\mathbf{s_{n-1}} + \mathbf{k_n}^\top \mathbf{v_n}, \\
&\mathrm{Ret}(\mathbf{X_{n}}) = \mathbf{q_n}\mathbf{s_n}, 
\end{aligned}
\end{equation}
where $\mathrm{Ret}$ and $\gamma$ denote retention and decay factor, respectively. In essence, $\mathbf{s_n}$ conveniently maintains the previous internal states. As shown in~\citep{sun2023retentive}, retention can also be defined in a parallel formulation
\begin{equation}
\mathrm{Ret}(\mathbf{X}) = (\mathbf{q}\mathbf{k}^\top 
 \odot \mathbf{M})\mathbf{v},
\end{equation}
Where $\mathrm{M}$ denotes a mask with a decay factor $\gamma$ as in
\begin{equation}
\mathbf{M_{ij}} =
\left\{
\begin{aligned}
& \gamma^{i-j}, &i\ge j \\
& 0, &i < j \\
\end{aligned}
\right.
\end{equation}
This dual representation of the retention in parallel and recurrent modes enables many desired properties, such as training parallelism and fast inference. For longer sequences, the recurrent mode can become inefficient. As a result, a hybrid approach, referred to as chunkwise, which combines recurrent and parallel formulation, is desired. Specifically, the input $\mathbf{X}$ is split into smaller sequences with chunksize $C$, in which $\mathbf{x}_{[m]} = [\mathbf{x}_{(m-1)C+1} , \cdots , \mathbf{x}_{mC}]$ represents the $m$-th chunk. The chunkwise query, key, and values can be defined as
\begin{equation}
\begin{aligned}
\label{eq:chunk1}
\quad \mathbf{q}_{[m]} = \mathbf{q}_{Cm:C(m+1)} &, \\
\quad \mathbf{k}_{[m]} = \mathbf{k}_{Cm:C(m+1)}&, \\
\quad \mathbf{v}_{[m]} = \mathbf{v}_{Cm:C(m+1)}&, \\
\end{aligned}
\end{equation}
The chunkwise retention formulation is as follows  
\begin{equation}
\begin{aligned}
\label{eq:ret:chunk}
\mathbf{R}_{m}&=\mathbf{k}_{[m]}^\top (\mathbf{v}_{[m]}\odot \zeta)+\gamma^{\mathbf{B}}\mathbf{R}_{m-1}, \\ 
\mathrm{Ret} (\mathbf{X}_{[m]}) &= (\mathbf{q}_{[m]} \mathbf{k}^\top_{[m]}\odot \mathbf{M}) \mathbf{v}_{[m]} +  (\mathbf{q}_{[m]}\mathbf{R}_{m-1}) \odot \xi, \\ \quad\xi_{mt}&=\gamma^{m+1}, \quad\zeta_{mt}=\gamma^{\mathbf{B}-m-1},& \\
\end{aligned}
\end{equation}


The underlying motivation of the chunkwise formulation is to employ the parallel mode in each chunk while processing cross-chunk representations in the recurrent mode. For high-resolution images with long sequences, the chunkwise formulation allows for faster processing of tokens and decoupling the memory. In Sec.~\ref{sec:throughput}, we demonstrate how ViRs compare more favorably to ViTs due to the chunkwise formulation for efficient processing of longer sequences.
\subsection{2D Retention}
We further expand the 1D formulation to achieve shift equivariance. Under 1D formulation, the decay between successive patches along a column of the image is increased by a factor of $W$ which is the number of patches per-row in the image. Our 2D formulation aims to maintain the decay between successive horizontal and vertical positions. 
\subsubsection{2D Recurrent Formulation}

Given a point $(x,y)$, we rewrite Eq.~\ref{eq:rnn} in the functional form $r(x,y)$ in order to parameterize the position within the sequence with both $x$ and $y$ coordinates, with $x,y \in \mathbb{Z}+$. We formulate it as
\begin{equation}
\begin{aligned}
    r(x+f,y) &= ... + \gamma^{f} r(x,y) + ... \\
    r(x,y+g) &= ... + \gamma^{g} r(x,y) + ...
\end{aligned}
\label{eq:equivariant_gamma}
\end{equation}
We adopt the L1 distance between position $(x+f,y+g)$ and $(x,y)$ as the decay rate which results in
\begin{equation}
    r(x+f,y+g) = ... + \gamma^{(x-f+y-g)} r(x,y) + ...
\end{equation}
We preserve the autoregressive property of retention, thus enforcing that $f, g \ge 0$. Furthermore, we derive the formulation of 2D retention in the recurrent form as in the following
\begin{equation}
\begin{aligned}
    r(1,1) & = \mathbf{k}_{1,1}^\intercal \mathbf{v}_{1,1}^{} \\
    r(x,1) & = \gamma r(x-1,1) + \mathbf{k}_{x,1}^\intercal \mathbf{v}_{x,1}^{} \\
    r(1,y) & = \gamma r(1,y-1) + \mathbf{k}_{1,y}^\intercal \mathbf{v}_{1,y}^{} \\
    r(x,y) & = \gamma r(x-1,y) + \gamma r(x,y-1) \\
           & - \gamma^2 r(x-1,y-1) + \mathbf{k}_{x,y}^\intercal \mathbf{v}_{x,y}^{}
\end{aligned}
\label{eq:2d_retention_recurrent}
\end{equation}
The first 3 terms of equation \ref{eq:2d_retention_recurrent} can be seen as base cases in the recursion. In fact, $r(x,1)$ and $r(1,y)$ take on the identical form of the original retention formulation. The intuition behind the generalized form $r(x,y)$ will become clearer in the next section when we introduce the parallel form of 2D Retention. Crucially, this form still allows for computing $r(x,y)$ with constant time complexity as is computes a sum over a fixed number of terms ($r(x-1,y)$, $r(x,y-1)$, $r(x-1,y-1)$). 



\subsubsection{2D Parallel Formulation}

For the convenience of notation, let $\Delta x = x - f$ and $\Delta y = y - g$ for some $f \le x$ and $g \le y$, and $x, y, f, g \in \mathbb{Z}^+$. Given this, we introduce the parallel formulation:

\begin{equation}
    p\left(x, y\right) = \sum_{g=1}^{y} \sum_{f=1}^{x} \gamma ^ {\left(\Delta x + \Delta y\right)} \mathbf{k}_{f,g}^\intercal \mathbf{v}_{f,g}^{}
\label{eq:2d_retention_parallel}
\end{equation}

It is also more apparent how the L1 distance underpins the decay rate as it is directly applied in the parallel formulation. Please see the supplementary material on the proof of equivalency between parallel and recurrent formulations.

To construct the full decay mask for the parallel formulation, we introduce the complete sequence of tokens $s \in S$, and position within, and then $x'(s) = s \mod W$ and $y'(s) = \left\lfloor s / W \right\rfloor$. Hence, $\Delta x' = x'(c) - x'(r)$ and $\Delta y' = y'(c) -y'(r)$. As a result, the mask is represented as

\begin{equation}
    \mathbf{M_{rc}} = \left\{
        \begin{aligned}
            & \gamma^{\left(\Delta x' + \Delta y' \right)}, & \Delta x' \ge 0, \Delta y' \ge 0 \\
            & 0, & \text{otherwise}
        \end{aligned}
    \right.
\end{equation}

\subsection{ViR Model}
\label{sec:model_d}
In the following, we first discuss the isotropic ViR model. In addition, we present the hybrid ViR, consisting of CNN and retention-based layers, which incorporates inductive biases such as locality and weight sharing that can improve training and data efficiency.

\subsection{Isotropic}
\label{sec:iso}
Fig.~\ref{fig:vir_arch} illustrates an overview of our proposed model. Given an input image $\mathbf{X} \in \mathbb{R}^{H \times W \times C}$ with height $H$ and width $W$, it is partitioned into patches and flattened into a sequence of tokens. This is similar to the tokenization scheme which was previously proposed by ViT~\citep{dosovitskiy2020image}. The tokenized patches are then projected into a patch embedding $Z = [\vz_1, \cdots, \vz_{|z|}] \in \mathbb{R}^{|z|\times D}$ with dimension $D$. Different from ViT, we first add the position embedding to the patch embedding and then append a \verb|[class]| token ($\mathbf{Z}_n^0=\mathbf{X}_\text{class}$). 

The output of the ViR encoder with $L$ layers ($\mathbf{Z}^n_L$) is used in a classification Mult-Layer Perceptron (MLP) head during both pre-training and finetuning. Due to the autoregressive nature of the ViR model, the position of the \verb|[class]| plays an important role as appending to the end of the embedding sequence acts as a summarizing of all the previous tokens. 

In lieu of self-attention, we use retention to enforce a recurrent formulation via masking. However, our formulation does not depend on gated retention or specific relative position embeddings (\textit{e.g.} xPos~\citep{sun2022length} or RoPE~\citep{su2021roformer}) and achieves numerical equivalency between parallel, recurrent and hybrid (\textit{i.e.} mixture of local recurrent and global parallel) formulations. Specifically, the parallel retention formulation solely depends on query $\mathbf{q}$, key $\mathbf{k}$, value $\mathbf{v}$ and a decay Mask $M$ and defined according to
\begin{equation}
\mathbf{q}, \mathbf{k},\mathbf{v} = \mathbf{z} \mathbf{A}_{qkv},
\end{equation}
\begin{equation}
\mathrm{Ret}(\mathbf{z}) = (\dfrac{\mathbf{q}\mathbf{k}^\top}{\sqrt{D_h}} \odot \mathbf{M})\mathbf{v},
\end{equation}
where $\mathrm{Ret}$ represents retention and $D_h$ is a scaling factor to balance the compute and parameter counts. In addition, the original retention formulation, as proposed in RetNet~\citep{sun2023retentive}, increases the number of parameters due to the addition of the learnable gated function, and a result decreases the image throughput under the same network layout. 


The retention ($\mathrm{Ret}$) is further extended to Multi-Head Retention (MHR). The retention is computed across each head with a constant decay factor and normalized with LayerNorm~\citep{layernorm} (LN) according to
\begin{equation}
\mathbf{Y}  =\mathrm{LN}([{\mathrm{Ret}}_1(\mathbf{z}); {\mathrm{Ret}}_2(\mathbf{z}); \cdots {\mathrm{Ret}}_k(\mathbf{z})]).
\end{equation}


We use alternating MHR and MLP blocks with LayerNorm (LN) and residual connections as the building blocks of the encoder according to
\begin{equation}
\begin{aligned}
\label{eq:arch}
\mathbf{Z^\prime}^{l} &= \mathrm{MHR}(\mathrm{LN}(\mathbf{Z}^{l})) + \mathbf{Z}^{l-1}, \\
\mathbf{Z}^{l} &= \mathrm{MLP}(\mathrm{LN}(\mathbf{Z^\prime}^{l})) + \mathbf{Z^\prime}^{l}.
\end{aligned}
\end{equation}
\subsection{Hybrid}
The Hybrid ViR (HViR) has a multi-scale architecture with four stages with different resolutions. The higher-resolution features are processed in the first two stages that comprise CNN-based blocks with residual connections. Specifically, given an input $\mathbf{h})$, it is defined as
\begin{align}
\begin{split}
\mathbf{\hat{h}} & = \text{GELU}(\text{BN}(\text{Conv}_{3\times3}(\mathbf{h}))), \\
\mathbf{h} & = \text{BN}(\text{Conv}_{3\times3}(\mathbf{\hat{h}})) + \mathbf{h}
\label{eq:fused_convmb}
\end{split}
\end{align}
Where $\text{Conv}_{3\times3}$ is a dense $3\times3$ convolutional layer and $\text{BN}$ denotes batch normalization~\cite{ioffe2015batch}. The lower resolution stages comprise of similar retention blocks as described in Sec.~\ref{sec:iso}. Please see the supplementary materials for architecture details of different HViR model variants. 

\begin{table}[!t]
\renewcommand\arraystretch{.7}
\centering
\tablestyle{6pt}{1.00}
\caption{Comparison of classification benchmarks on \textbf{ImageNet-1K} dataset~\cite{deng2009imagenet}. Image throughput is measured on A100 GPUs with a batch size of 128. Models with $^\ddag$ are pre-trained on ImageNet-21K dataset.} 
\vspace{-1mm}
\resizebox{0.96\linewidth}{!}{
\setlength{\tabcolsep}{.5mm}{
\begin{tabular}[t]{lccccc}
\toprule
Model &  Image Size& \#Param & FLOPs & Throughput& Top-1 \\
      &  (Px) & (M) & (G) & (Img/Sec) & (\%) \\
\midrule
\multicolumn{6}{c}{Conv-Based} \\
\midrule
ConvNeXt-T~\cite{liu2022convnet} & 224 &\ \  28.6 &\ \ 4.5 & 3196 & 82.0\\
ConvNeXt-S~\cite{liu2022convnet} & 224 &\ \  50.2 &\ \ 8.7 & 2008 & 83.1\\
ConvNeXt-B~\cite{liu2022convnet} & 224 &\ \  88.6 & 15.4 & 1485 & 83.8\\
RegNetY-040~\cite{radosavovic2020designing} & 288 &\ \  20.6 &\ \  6.6 & 3227 & 83.0\\
ResNetV2-101~\cite{wightman2021resnet}& 224 &\ \  44.5 &\ \  7.8 & 4019 & 82.0\\
EfficientNetV2-S~\cite{tan2021efficientnetv2} & 384 &\ \  21.5 &\ \  8.0 & 1735 & 83.9\\
\midrule
\multicolumn{6}{c}{Transformer-Based} \\
\midrule
Swin-T~\cite{liu2021swin} & 224 &\ \  28.3 &\ \  4.4 & 2758 & 81.3\\
Swin-S~\cite{liu2021swin} & 224 &\ \  49.6 &\ \  8.5 & 1720 & 83.2\\
SwinV2-T~\cite{liu2022swin} & 256 &\ \  28.3 &\ \  4.4 & 1674 & 81.8\\
SwinV2-S~\cite{liu2022swin} & 256 &\ \  49.7 &\ \  8.5 & 1043 & 83.8\\
SwinV2-B~\cite{liu2022swin} & 256 &\ \  87.9 & 15.1 &\ \  535 & 84.6\\
Twins-S~\cite{chu2021twins} & 224 &\ \  24.1 &\ \  2.8 & 3596 & 81.7\\
Twins-B~\cite{chu2021twins} & 224 &\ \  56.1 &\ \  8.3 & 1926 & 83.1\\
Twins-L~\cite{chu2021twins} & 224 &\ \  99.3 & 14.8 & 1439 & 83.7\\
DeiT-S~\cite{touvron2021training} & 224 & 22.1 & 4.2 & 4608 & 79.9\\
DeiT-B~\cite{touvron2021training} & 224 &\ \  86.6 & 16.9 & 2035 & 82.0\\
DeiT-B~\cite{touvron2021training} & 384 &\ \  86.9 & 49.4 &\ \  480 & 83.1\\
DeiT3-B & 224 & 86.6 & 16.9 & 670 & 83.8 \\
DeiT3-L & 224 & 304.4 & 59.7 &\ \ 535  & 84.8	\\
PoolFormer-S36~\cite{yu2022metaformer} & 224 & 30.9 & 5.0 & 1656 & 81.4\\
PoolFormer-M36~\cite{yu2022metaformer} & 224 & 56.2 & 8.8 & 1170 & 82.1\\ 
PoolFormer-M58~\cite{yu2022metaformer} & 224 &\ \  73.5 & 11.6 &\ \  884 & 82.4\\
\midrule
\multicolumn{6}{c}{Hybrid} \\
\midrule
CoaT-Lite-S~\cite{xu2021co} & 224 &\ \  19.8 & \ \  4.1 & 2269 & 82.3\\
CrossViT-S~\cite{chen2021crossvit} & 240 & 26.9 & 5.1 & 2832 & 81.0\\
CrossViT-B~\cite{chen2021crossvit} & 240 & 105.0 & 20.1 & 1321 & 82.2\\
Visformer-S~\cite{chen2021visformer} & 224 &\ \  40.2 &\ \  4.8 & 3676 & 82.1\\

EdgeViT-S~\cite{edgevit} & 224 &\ \  13.1 & \ \ 1.9 & 4254 & 81.0\\ 
EfficientFormer-L7~\cite{li2022efficientformer} & 224 &\ \  82.2 & 10.2 & 1359 & 83.4\\
MaxViT-B~\cite{tu2022maxvit} & 224 &\ \  120.0 & \ \ 23.4 & 507 & 84.9\\
MaxViT-L~\cite{tu2022maxvit} & 224 &\ \  212.0 & \ \ 43.9 & 376 & 85.1\\



\midrule
\multicolumn{6}{c}{\textbf{ViR}} \\
\midrule
\rowcolor{Gray} 
\rowcolor{Gray}
\textbf{HViR-0} &  224 & 23.7 & 3.2 & \textbf{5012} &  \textbf{81.1} \\
\rowcolor{Gray}
\textbf{HViR-1} &  224 & 39.1 & 4.9 & \textbf{3680} &  \textbf{82.6} \\
\rowcolor{Gray}
\textbf{HViR-2} &  224 & 56.5 & 8.2 & \textbf{2820} &  \textbf{83.3} \\
\rowcolor{Gray}
\textbf{HViR-3} &  224 & 112.6 & 17.0 & \textbf{1510} &  \textbf{84.6} \\
\rowcolor{Gray}
\textbf{HViR-4} &  224 & 397.2 & 26.3 & \textbf{650} &  \textbf{84.8} \\
\rowcolor{Gray}
\textbf{ViR-S/16} & 224 & \ \ 31.4 &\ \  3.3 & \textbf{1621} &  \textbf{81.0} \\
\rowcolor{Gray} 
\textbf{ViR-B/16} &  224 &\ \  53.4 &\ \  5.3 & \textbf{671} &  \textbf{82.6}\\
\rowcolor{Gray} 
\textbf{ViR-L/16} &  224 &\ \  304.4 &\ \  59.7 & \textbf{531} &  \textbf{83.7} \\
\rowcolor{Gray}
\textbf{ViR-L/14}$^\ddag$ &  224 &\ \  304.4 &\ \  77.8 & \textbf{429} &  \textbf{85.0} \\
\rowcolor{Gray}
\textbf{ViR-L/14}$^\ddag$ &  448 &\ \  304.4 &\ \  310.3 & \textbf{319} &  \textbf{86.1} \\

\bottomrule
\end{tabular}}
}
\label{tab:imagenet}
\end{table}

\section{Experiments}


\subsection{Classification}
We present image classification benchmarks on ImageNet-1K dataset ~\citep{deng2009imagenet} in Table~\ref{tab:imagenet}. The ViR models demonstrate competitive performance across different model variants. Specifically, both ViR and HViR variants compare favorably to ViT-based counterparts, considering the Top-1 accuracy and image throughput tradeoff. For larger models, the ViR-L/14 model also achieves competitive performance when pretrained and finetuned on ImageNet-21K and ImageNet-1K datasets, respectively. In addition, Increasing the image resolution from $224 \times 224$ to $448 \times 448$ during the finetuning results in a considerable +1.1\% improvement in terms of Top-1 accuracy. Hence, it validates the effectiveness and scalability of ViR models for training on larger datasets and higher image resolutions. 

\subsection{Downstream Tasks}
In Table~\ref{tab:cascademaskrcnn}, we present object detection and instance segmentation benchmarks on MS COCO dataset~\cite{lin2014microsoft} for models that use Cascade Mask R-CNN~\cite{he2017mask} head. Models with HViR backbones compare favorably and outperform ConvNeXt~\cite{liu2022convnet} and Swin~\cite{liu2021swin} counterparts by +1.3 and +3.7 for HViR-1 and +0.2 and +0.2 for HViR-2 in terms of box AP, respectively. In addition, we present semantic segmentation benchmarks on ADE20K dataset~\cite{zhou2017scene} for models with UPerNet~\cite{xiao2018unified} segmentation head in Table~\ref{tab:cascademaskrcnn}. Models with HViR backbones outperform ConvNeXt and Swin counterparts by +0.3 and +2.5 for HViR-1 and +0.1 and +2.1 in terms of mIoU, respectively.

\begin{table}[!t]
    \centering
    \caption{Benchmarks for object detection and instance segmentation experiments on \textbf{MS COCO} dataset~\cite{lin2014microsoft}. Cascade Mask R-CNN~\cite{he2017mask} is used as a detection head. All models use a $3\times$ schedule. Statistics are computed using an input test resolution of $1280\times 800$.}
    \label{tab:cascademaskrcnn}
    \resizebox{1.0\linewidth}{!}{
    \begin{tabular}{lc|ccc|ccc}
        \toprule 
        Backbone & Throu. &  \multicolumn{3}{c|}{$\text{AP}^{\text{box}}$}  & \multicolumn{3}{c}{$\text{AP}^{\text{mask}}$} \\ 
         & im/sec &  Box & 50 & 75 & Mask & 50 & 75  \\ 
        \midrule
  DeiT-S/16~\cite{touvron2021training} & 269 &  48.0 & 67.2 & 51.7 & 41.4 & 64.2 & 44.3 \\
        Swin-T~\cite{liu2021swin} & 161 &  50.4 & 69.2 & 54.7 & 43.7 & 66.6 & 47.3 \\
        ConvNeXt-T~\cite{liu2022convnet} & 166 &  50.4 & 69.1 & 54.8 & 43.7 & 66.5 & 47.3 \\

        \rowcolor{Gray}
        \textbf{HViR-1} & \textbf{274} &  \textbf{51.7} & \textbf{69.8} & \textbf{55.3} & \textbf{44.1} & \textbf{67.3} & \textbf{48.2}  \\
        
        \midrule
        
        
        Swin-S~\cite{liu2021swin} & 119 &  51.9 & 70.7 & 56.3 & 45.0 & 68.2 & 48.8 \\
        		X101-32~\cite{xie2017aggregated} & 124 &  48.1 & 66.5 & 52.4 & 41.6 & 63.9 & 45.2 \\
        ConvNeXt-S~\cite{liu2022convnet} & 128 &  51.9 & 70.8 & 56.5 & 45.0 & 68.4 & 49.1 \\
        \rowcolor{Gray}
        \textbf{HViR-2} & \textbf{138} &  \textbf{52.1} & \textbf{71.0} & \textbf{56.6} & \textbf{45.2} & \textbf{68.5} & \textbf{49.2} \\
        \bottomrule
    \end{tabular}
    }
\end{table}

\begin{table}[t]
\centering
\caption{Benchmarks for semantic segmentation experiments on \textbf{ADE20K} dataset~\cite{zhou2017scene} using UPerNet~\cite{xiao2018unified} network. Throughput is reported in image/sec by using an input test resolution of $512\times 512$.}
\label{tab:adek20kseg}
\resizebox{1\linewidth}{!}{
\footnotesize
  \begin{tabular}{lcccc}
    \toprule
    Model  & Throughput & FLOPs (G) & mIoU \\
    \midrule	 
    Swin-T~\cite{liu2021swin}  & 350 & 945 & 44.5\\
    ConvNeXt-T~\cite{liu2022convnet}  & 363 & 939  & 46.7\\
         \rowcolor{Gray}
    \textbf{HViR-1}  & \textbf{371} & 958  & \textbf{47.0}\\
    \midrule
    Swin-S~\cite{liu2021swin}  & 219 & 1038  & 47.6\\
    Twins-SVT-B~\cite{chu2021twins}  & 204 &  - & 47.7\\
    ConvNeXt-S~\cite{liu2022convnet}  & 234 & 1027  &49.6\\
        \rowcolor{Gray}
    \textbf{HViR-2}  & \textbf{241} & 1041  & \textbf{49.7}\\
    \bottomrule

  \end{tabular} 
  }
    \label{tab:ade_segmentation}
\end{table}

\begin{figure}
    \centering
    \includegraphics[width=1\linewidth]{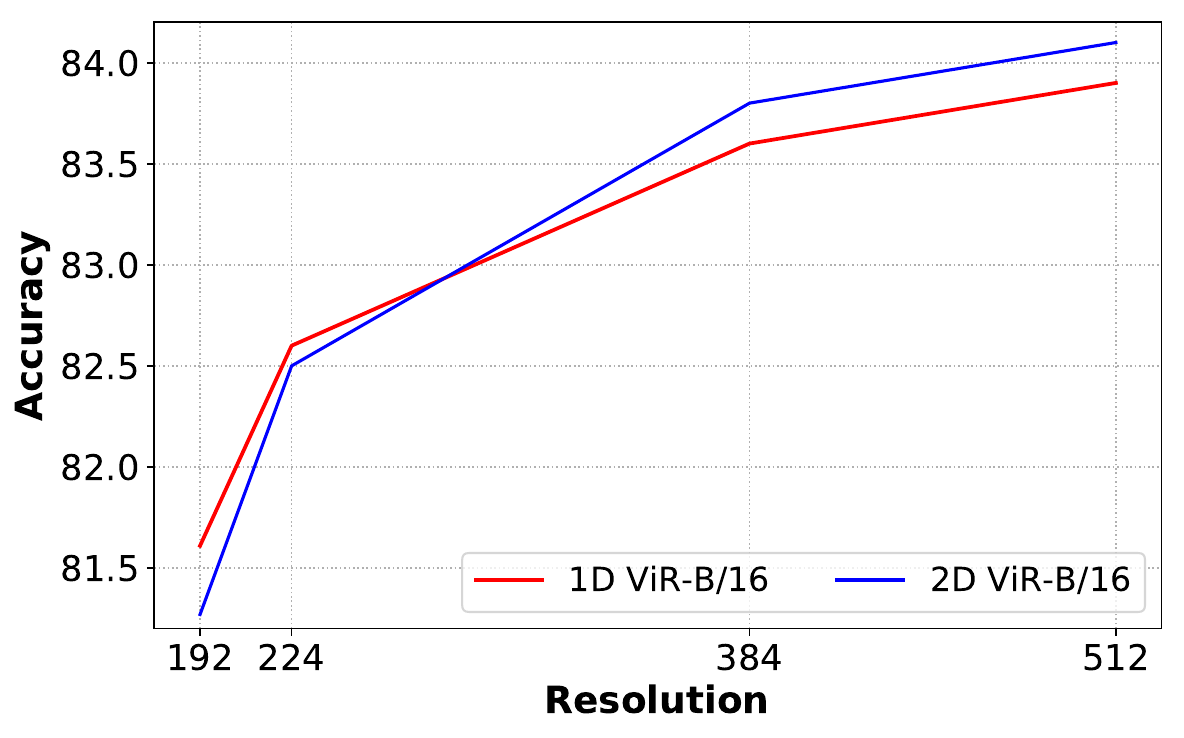}
    \caption{Effect of increasing the image resolution on Top-1 accuracy for 1D ViR-B/16 and 2D ViR-B/16 networks. }
    \label{fig:1d_2d_comparison_v2}
\end{figure}

\begin{figure}[]
\centering

\resizebox{1.\linewidth}{!}{
\begingroup
\renewcommand*{\arraystretch}{0.3}
\begin{tabular}{c}

  \includegraphics[width=1\linewidth]{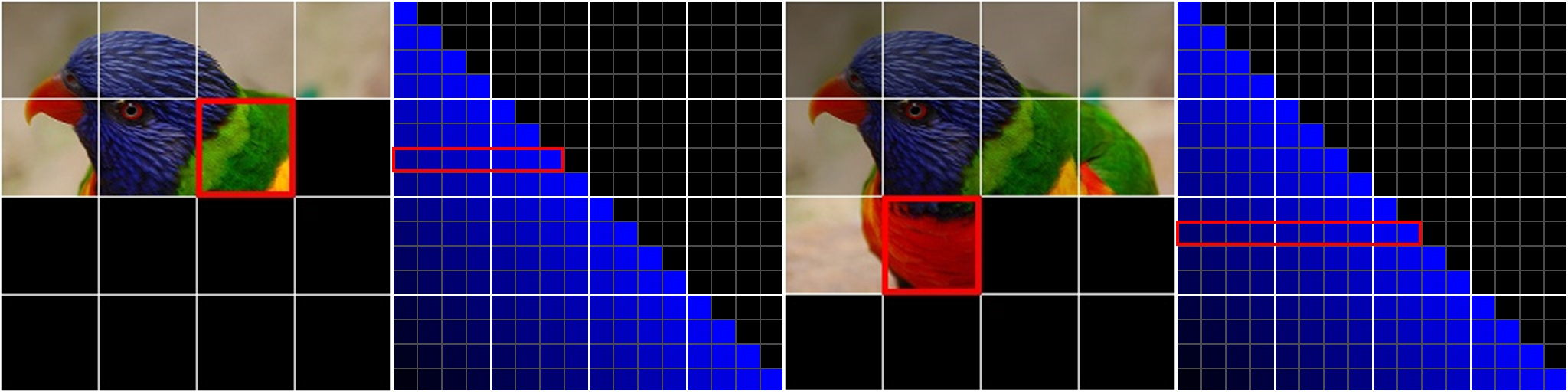}\vspace{1mm} \\ 
  (a) 1D Retention.\vspace{1mm} \\
   \includegraphics[width=1\linewidth]{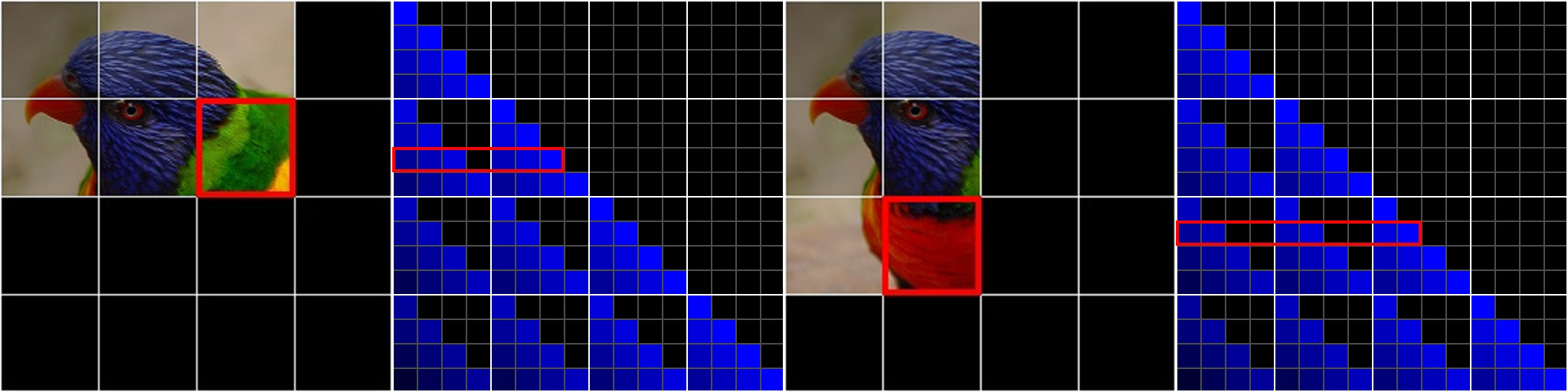}\vspace{1mm} \\ 
  (b) 2D Retention. \\
\end{tabular}
\endgroup
}
\caption{Effective receptive field and corresponding masks for: (a)1D retention (b) 2D retention. Cell opacity is based on the decay strength given the distance between the highlighted cell and each of the colored in cells. The 2D formulation achieves shift equivariance, enabling an identical decay factor between successive horizontal and vertical positions. Hence 2D retention is more suitable for finetuning on higher resolutions.}
\label{fig:1d_2d_comparison}
\end{figure}

\begin{table}
\resizebox{1\linewidth}{!}{
\centering
\footnotesize
\setlength{\tabcolsep}{2.5pt}
\begin{tabular}{l|c|cc|c}
\Xhline{1.0pt}
 & \multicolumn{1}{c|}{ImageNet} & \multicolumn{2}{c|}{COCO} & \multicolumn{1}{c}{ADE20k} \\
 & Top-1   & AP$^\text{box}$ & AP$^\text{mask}$ & mIoU \\
\hline
HViR-1 (1D Retention) & \textbf{82.3} & 51.2 & 43.8 & 46.9 \\
HViR-1 (2D Retention) & 82.2 & \textbf{51.7} & \textbf{44.1} & \textbf{47.0} \\
\Xhline{1.0pt}
\end{tabular}
  }
    \caption{Ablation study on the effectiveness of 1D and 2D formulations for different tasks with HViR-1 model.}
    \label{tab:abl_study_cmt}
\end{table}

\begin{figure}
    \centering
    \includegraphics[width=1\linewidth]{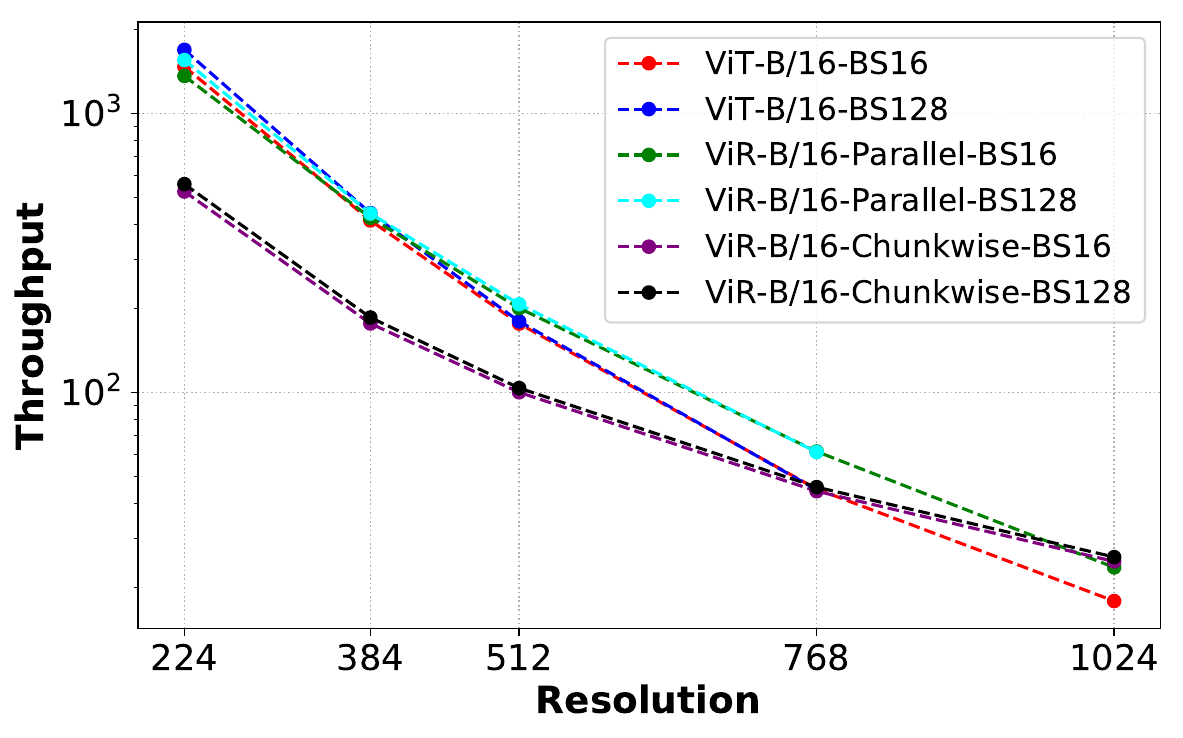}
    \caption{Effect of image resolution on throughput for ViR-B/16 and ViT-B/16 models. Throughput is measured on an A100 80GB NVIDIA GPU with batch sizes of 16 and 128. For a batch size of 128, the memory is insufficient to process images for both ViT and parallel mode of ViR networks. For $1024 \times 1024$, ViR-B/16 with chunkwise mode is the only configuration that can process images with batch size of 128.}
    \label{fig:throughput_b}
\end{figure}

\begin{figure*}[t]
\centering

\resizebox{0.85\linewidth}{!}{
\begingroup
\renewcommand*{\arraystretch}{0.3}
\begin{tabular}{c}

  \includegraphics[width=1\linewidth]{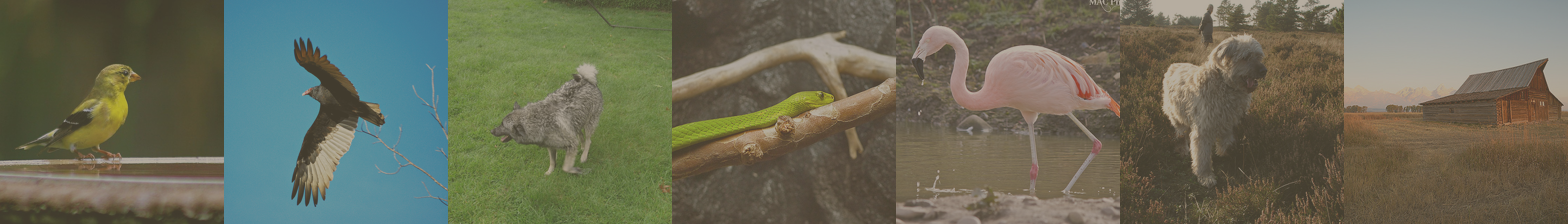} \\ 
  {(a) Images from ImageNet-1K validation set.} \\[3pt]
 
\includegraphics[width=1\linewidth]{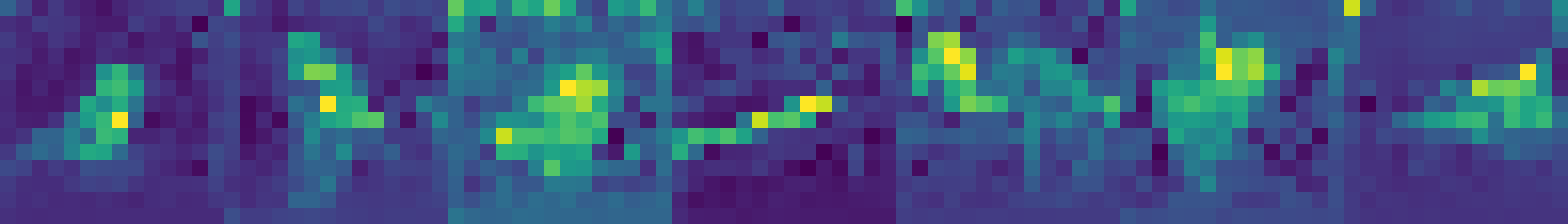} \\
{(b) \textbf{Retention} maps.}

\end{tabular}
\endgroup
}
\caption{Visualization of : (a) input images (b) retention maps. Salient image features are localized in the retention maps. In addition, both short and long-range spatial dependencies have been captured effectively.} 
\vspace{-2mm}
\label{fig:retention_maps}
\end{figure*}

\section{Ablation}
\label{sec:abl}

\subsection{Component Study}
\label{sec:comp_study}
In this section, we study the effect of different component design choices on the overall performance by examining the Top-1 and throughput trade-offs. As the base model, we use a 1D ViR-B/16 with a Top-1 accuracy of 82.6\% on ImageNet-1K dataset. First, we investigate the effect of \verb|[class]| token by removing it and using a global average pooling layer before the classification head. In this case, the Top-1 accuracy decreases by 0.4\%. As discussed in Sec.\ref{sec:model_d}, the \verb|[class]| plays an important role as it encapsulates global information from the preceding tokens that are useful for the task of image classification. In addition, this change reduces the throughput by 4.91\%. We also study the effect of adding a gated function to the retention mechanism. For fair comparison, we reduced the number of layers to match the same number of parameters as the base model. However, this configuration decreased the image throughput and Top-1 accuracy by 7.45\% and 0.5\% respectively. We also investigated the effect of scaling the key tensor, in the lie of the query. Although image throughput and Top-1 accuracy remain roughly unchanged, we observed some instabilities with sudden changes in loss values during training. In addition, as opposed to an autoregressive formulation, we also investigated the use of multipass encoding by providing both left and right token orders. Our results show that although Top-1 accuracy is improved by +0.3\%, the throughput is significantly impacted and reduced by +51.11\%. Hence, multipass encoding does not provide an optimal performance vs.\ efficiency tradeoff in this case. 
\begin{table}
\centering
\resizebox{.85\linewidth}{!}{
\setlength{\tabcolsep}{2.5pt}
  \begin{tabular}{lcc}
    \toprule
    Design Component & Throughput (im/sec)   & Top-1 (\%) \\
    \midrule	 
    Gated retention  & 621& 82.1 \\
    No class token  & 638& 82.2 \\
    Key ($\mathbf{k}$) scaling  & 665& 82.5 \\
     Multipass encoding &328  & 82.9 \\
    \rowcolor{Gray}
    \textbf{Base Model}& 671  & 82.6\\
    \bottomrule
  \end{tabular} 
  }
    \caption{Ablation study on the effect of different design choices on ImageNet Top-1 accuracy vs throughput performance tradeoff. The throughput is measured on an A100 80GB NVIDIA GPU with a batch size of 128. The base model is ViR-B/16.}
    \label{tab:abl-sup-downsampler}
\end{table}

\subsection{1D and 2D Retention Comparison}
\paragraph{Higher Resolution Finetuning}
In Fig.~\ref{fig:1d_2d_comparison_v2}, we illustrate the effect of scaling the image resolution on the Top-1 accuracy of 1D and 2D ViR-B/16 models. For each variant, a base model has been trained on $192\times192$ resolution and fine-tuned on various higher image sizes. Specifically, 2D ViR-B/16 model shows better performance in comparison to its 1D counterpart due to its desirable shift equivariance property which maintains an identical decay factor between successive patches in vertical and horizontal directions. 
\paragraph{Propagation Pattern}
In addition, in Fig.~\ref{fig:1d_2d_comparison}, we demonstrate a qualitative comparison between 1D and 2D retention mechanisms by showing the relationship between a patch (\textit{i.e.} red border) and other patches in its receptive field. Due to the auto-regressive nature of retention, we can see how the receptive field can only attend to previously encountered patches within the image. Additionally, the strength of the connection between two patches is decayed based on the distance between them. Since we read out images as scanlines, the distance is based on the number of patches processed and not on any concept of two-dimensional distance.
\paragraph{Downstream Tasks}
In Table~\ref{tab:abl-sup-downsampler}, we present quantitative comparisons for the performance of HViR-1 model with 1D and 2D retention formulations across different tasks. For ImageNet classification, model with 1D formulation slightly outperforms the 2D counterpart. However, 2D retention outperforms the model with 1D formulation on object detection and instance segmentation by +0.5 and +0.3 in terms of box AP and mask AP, respectively and +0.1 in terms of mIoU for semantic segmentation. Hence, these benchmarks demonstrate the effectiveness of 2D retention formulation for downstream tasks with higher resolution images.

\subsection{Computational Analysis}
\label{sec:throughput}

\paragraph{Complexity}
The primary motivation behind ViR is to find a formulation that allows for high inference throughput without sacrificing model quality. Given an input image with height $H$ and width $W$ and a patch size of $P$ and a sequence length of $N=\frac{HW}{P^2}$, a regular attention mechanism has a complexity of $O\left(\frac{H^2W^2}{P^4}\right)$ which significantly impacts the efficiency for higher resolution images. In ViR, since the recurrent formulation only depends on the previous token for next token prediction, the complexity with respect to the input is of $O(N)$. Although the Parallel mode can process all tokens simultaneously this comes with the quadratic scaling complexity of $O(N^2)$. The chunkwise mode is combination of parallel and recurrent modes in which each chunk only depends on the previous one. Within each chunk, the parallel mode is used. Specifically, given a chunk size $C$ and sequence length $N$, the number of chunks is $\ceil{\frac{N}{C}}$ and per-chunk complexity is $O(C^2)$. Hence, the overall complexity is of $O(NC)$. 

\paragraph{Memory}
In addition to throughput improvements, recurrent and chunkwise also adopt desirable memory properties. For downstream applications without patch-based features (\textit{e.g.} image classification), the memory complexities for recurrent and chunkwise formulations are $O(1)$ and $O(C^2)$, respectively for a chunk size of $O(C)$. For application that require patch-based features, the memory complexities are $O(N)$ and $O(N + C^2)$, respectively. Fig.~\ref{fig:throughput_b} shows the impact of input image size on throughput for ViR-B/16 and ViT-B/16 models. Specifically, ViR demonstrates favorable scaling characteristics as the resolution increases. At very high resolutions, only ViR-B/16 with chunkwise formulation can process images on an A100 80GB NVIDIA GPU. In this case, the memory is insufficient for ViT-B/16. In addition, due to the different compute complexity scaling rules between parallel and chunkwise formulations, the chunkwise shows higher image throughput than counterpart parallel at very high resolutions. Our analysis shows similar findings for ViR-L/16 and ViT-L/16 models. Please refer to the supplementary materials for more details.   

\subsection{What Does Retention See?}
\label{sec:retention_maps}
In Fig.~\ref{fig:retention_maps}, we illustrate retention maps that are obtained from an ImageNet-1K pretrained ViR-S/16 model. Specifically, the retention maps are extracted from the last layer of the encoder without using any post-processing or normalization layers. We observe that high-intensity response regions correspond to salient image features. For elongated objects, the long-range spatial dependencies have been effectively captured. We observe similar trends in other ViR variants that are trained on both ImageNet-1K and ImageNet-21K datasets.

\section{Conclusion}
In this work, we introduced a new class of computer vision models, referred to as Vision Retention Networks (ViR), with dual parallel and recurrent formulations. The equivalency of these formulations allows for desired properties such as training parallelism and fast inference while maintaining great performance. In addition, a hybrid formulation, denoted as chunkwise, enables the processing of longer sequences with considerably more efficient time and space complexities. We have trained and tested the proposed ViR on ImageNet-1K and ImageNet-21K datasets with different resolutions and achieved competitive performance. We believe the proposed ViR could be the foundation of a new class of efficient vision-friendly models that offer training and inference flexibility for a variety of applications.

{
    \small
    \bibliographystyle{ieeenat_fullname}
    \bibliography{main}
}

\onecolumn
\renewcommand{\thefootnote}{\arabic{footnote}}
\renewcommand{\thesection}{\Alph{section}}
\renewcommand\thefigure{S.\arabic{figure}}
\setcounter{figure}{0}
\renewcommand\thetable{S.\arabic{table}}
\setcounter{table}{0}

\appendix

\section{ViR-2D Mathematical Formulation}\label{apdx:2d_proof}

\subsection{Proof of Recurrent and Parallel Equivalence}\label{appendix:identity_proof}

First, as a reminder, these are the two forms of our decay formulation. To simplify notation, let $\mathbf{z}_{mn} := \mathbf{k}_{m,n}^\intercal \mathbf{v}_{m,n}$. Also, we use $r(x,y)$ and $p(x,y)$ to distinguish the recurrent and parallel forms respectively.

\paragraph{Recurrent Form}

\begin{equation}
\begin{split}
    r(1,1) & = \mathbf{z}_{11} \\
    r(x,1) & = \gamma r(x-1,1) + \mathbf{z}_{x1} \\
    r(1,y) & = \gamma r(1,y-1) + \mathbf{z}_{1y} \\
    r(x,y) & = \gamma r(x-1,y) + \gamma r(x,y-1) - \gamma^2 r(x-1,y-1) + \mathbf{z}_{xy}
\end{split}
\tag{\ref{eq:2d_retention_recurrent} revisited}
\end{equation}

\paragraph{Parallel Form}

\begin{equation}
    p\left(x, y\right) = \sum_{g=1}^{y} \sum_{f=1}^{x} \gamma ^ {\left(\Delta x + \Delta y\right)} \mathbf{z}_{fg}
\tag{\ref{eq:2d_retention_parallel} revisited}
\end{equation}

with $\Delta x = x - f$ and $\Delta y = y - g$.

\paragraph{Proof} 
First, we show the equivalence of the three special cases of the recurrent form:

\subsubsection{Cell $(1,1)$}\label{proof:cell11}

For cell $(1,1)$ we have:

\begin{equation}
\begin{split}
    r(1,1) & = \mathbf{z}_{11} \\
    p(1,1) & = \sum_{g=1}^{1} \sum_{f=1}^{1} \gamma^{\Delta x + \Delta y} \mathbf{z}_{11} \\
           & = \gamma^{\left[(1-1)+(1-1)\right]} \mathbf{z}_{11} \\
           & = \gamma^0 \mathbf{z}_{11} \\
           & = \mathbf{z}_{11}
\end{split}
\label{eq:proof11}
\end{equation}

Thus $r(1,1) = p(1,1)$. QED.

\subsubsection{Row $(x,1)$}\label{proof:rowx1}

To simplify the parallel expression, when $g = y$, then $\Delta y = 0$, and $\sum_{i=1}^{1} h(..., i) = h(..., 1)$, then we get a parallel form for the first row of values as only being dependent on $x$.

\begin{equation}
\begin{split}
    p(x,1) & = \sum_{f=1}^x \gamma^{\Delta x} \mathbf{z}_{f1}^{} \\
           & = \sum_{f=1}^x \gamma^{x - f} \mathbf{z}_{f1}^{}
\end{split}
\label{eq:pl_first_row}
\end{equation}

We can see that:

\begin{equation}
\begin{split}
    p(x+1,1) & = \sum_{f=1}^{x+1} \gamma^{(x + 1) - f} \mathbf{z}_{f,1}^{} \\
             & = \sum_{f=1}^{x+1} \gamma \cdot \gamma^{x - f} \mathbf{z}_{f,1}^{} \\
             & = \gamma \sum_{f=1}^{x+1} \gamma^{x - f} \mathbf{z}_{f,1}^{} \\
             & = \gamma \left[ \left(\underbrace{\sum_{f=1}^x \gamma^{x - f} \mathbf{z}_{f1}^{}}_\text{Sum up to x} \right) + \underbrace{\gamma^{x-(x+1)} \mathbf{z}_{x+1,1}}_\text{Summand for $x+1$}  \right] \\
             & = \gamma \left[ \left(\sum_{f=1}^x \gamma^{x - f} \mathbf{z}_{f1}^{} \right) + \gamma^{-1} \mathbf{z}_{x+1,1}  \right] \\
             & = \gamma \left[ \sum_{f=1}^x \gamma^{x - f} \mathbf{z}_{f1}^{} \right] + \mathbf{z}_{x+1,1} \\
             & = \gamma \left[ p(x,1) \right] + \mathbf{z}_{x+1,1} \\
             & = \gamma p(x,1) + \mathbf{z}_{x+1,1}
\end{split}
\label{eq:proof_first_row_parallel}
\end{equation}

If we have that

\begin{equation}
\begin{split}
    r(x+1,1) &= \gamma r(x,1) + \mathbf{z}_{x+1,1} \\
    p(x+1,1) &= \gamma p(x,1) + \mathbf{z}_{x+1,1}
\end{split}
\end{equation}

and we've shown that $r(1,1) = p(1,1)$ in the proof for cell $(1,1)$, then because both $r(x+1,1)$ and $p(x+1,1)$ take on the same form, $r(x+1,1) = p(x+1,1)$. QED.

Incidentally, this also serves as a proof of equivalence for the 1D form of retention, which was left implicitly defined in \cite{sun2023retentive}.

\subsubsection{Column $(1,y)$}\label{proof:col1y}

Because $x$ and $y$ are treated independently in the parallel formulation, and we have that

\begin{equation}
    p(1,y) = \sum_{g=1}^y \gamma^{\Delta y} \mathbf{z}_{1,g}^{}
\end{equation}

We can see that the proof for the first column trivially follows that of the proof for the first row, and we leave the exercise up to the reader.

\subsubsection{Any Cell $(x,y)$ s.t. $x>1$ and $y>1$}

Given that we've shown equivalence for the first row and the first column, we can turn our attention to the general case. First, we write $p(x-1,y)$ in terms of $p(x,y)$.

\begin{equation}
\begin{split}
    p(x-1,y) & = \sum_{g=1}^{y} \sum_{f=1}^{x-1} \gamma^{\left(\Delta x - 1 + \Delta y \right)} \mathbf{z}_{fg}^{} \\
             & = \gamma^{-1} \sum_{g=1}^{y} \sum_{f=1}^{x-1} \gamma^{\left(\Delta x + \Delta y \right)} \mathbf{z}_{fg}^{} \\
             & = \gamma^{-1} \left[\underbrace{\left(\sum_{g=1}^{y} \sum_{f=1}^{x} \gamma^{\left(\Delta x + \Delta y \right)} \mathbf{z}_{fg}^{} \right)}_\text{sum to x} - \underbrace{\sum_{g=1}^{y} \gamma^{\Delta y} \mathbf{z}_{xg}^{}}_\text{subtract introduced f=x terms} \right] \\
             & = \gamma^{-1} p(x,y) - \sum_{g=1}^{y}  \gamma^{(\Delta y - 1)} \mathbf{z}_{xg}^{} \\
             \\
    p(x,y)   & = \gamma p(x - 1,y) + \sum_{g=1}^{y} \gamma^{\Delta y} \mathbf{z}_{xg}^{}
\end{split}
\label{eq:cellxp1y}
\end{equation}

This yields the following relationship between successive x coordinates:

\begin{equation}
    p(x,y) - \gamma p(x-1,y) = \sum_{g=1}^{y} \gamma^{\Delta y} \mathbf{z}_{xg}^{}
\label{eq:stepx}
\end{equation}

We'll use that final form later on in $p(x-1,y-1)$. The same steps also hold for $p(x,y-1)$:

\begin{equation}
\begin{split}
    p(x, y-1) & = \gamma^{-1} p(x, y) - \sum_{f=1}^x \gamma^{(\Delta x - 1)} \mathbf{z}_{fy}^{} \\
    \\
    p(x,y) & = \gamma p(x, y - 1) + \sum_{f=1}^x \gamma^{\Delta x} \mathbf{z}_{fy}^{}
\end{split}
\label{eq:pxym1}
\end{equation}

And again, the relation between successive y coordinates:

\begin{equation}
    p(x,y) - \gamma p(x,y-1) = \sum_{f=1}^x \gamma^{\Delta x} \mathbf{z}_{fy}^{}
\label{eq:stepy}
\end{equation}

Finally, we'll rewrite $p(x-1,y-1)$ in terms of $p(x,y)$:

\begin{equation}
\begin{split}
    p(x-1,y-1) & = \sum_{g=1}^{y-1} \sum_{f=1}^{x-1} \gamma^{(\Delta x - 1 + \Delta y - 1)} \mathbf{z}_{fg}^{} \\
               & = \gamma^{-2} \sum_{g=1}^{y-1} \sum_{f=1}^{x-1} \gamma^{(\Delta x + \Delta y)} \mathbf{z}_{fg}^{} \\
               & = \gamma^{-2} \sum_{g=1}^{y-1} \left[\underbrace{\left(\sum_{f=1}^x \gamma^{(\Delta x + \Delta y)} \mathbf{z}_{fg} \right) - \left( \gamma^{\Delta y} \mathbf{z}_{xg}^{} \right)}_\text{sum to x}  \right] \\
               & = \gamma^{-2} \left[\left(\sum_{g=1}^{y-1} \sum_{f=1}^x \gamma^{(\Delta x + \Delta y)} \mathbf{z}_{fg} \right) - \left(\sum_{g=1}^{y-1} \gamma^{\Delta y} \mathbf{z}_{xg} \right) \right] \\
    \gamma^2 p(x-1,y-1) & = \left[\underbrace{\left(\sum_{g=1}^{y} \sum_{f=1}^x \gamma^{(\Delta x + \Delta y)} \mathbf{z}_{fg} \right) - \left(\sum_{f=1}^x \gamma^{\Delta x} \mathbf{z}_{fy} \right)}_\text{sum to y} \right] - \left(\sum_{g=1}^{y-1} \gamma^{\Delta y} \mathbf{z}_{xg} \right) \\
    \gamma^2 p(x-1,y-1) & = p(x,y) - \left(\sum_{f=1}^x \gamma^{\Delta x} \mathbf{z}_{fy} \right) - \left(\sum_{g=1}^{y-1} \gamma^{\Delta y} \mathbf{z}_{xg} \right) \\
    \gamma^2 p(x-1,y-1) & = p(x,y) - \underbrace{\left[p(x,y) - \gamma p(x,y-1)\right]}_\text{Equation \ref{eq:stepy}} - \left(\sum_{g=1}^{y-1} \gamma^{\Delta y} \mathbf{z}_{xg} \right) \\
    \gamma^2 p(x-1,y-1) & = \gamma p(x,y-1) - \left[\underbrace{\left(\sum_{g=1}^{y} \gamma^{\Delta y} \mathbf{z}_{xg} \right) - \gamma^{0} \mathbf{z}_{xy}^{}}_\text{sum to y} \right] \\
    \gamma^2 p(x-1,y-1) & = \gamma p(x,y-1) - \left[\underbrace{\left(p(x,y) - \gamma p(x-1,y)\right)}_\text{Equation \ref{eq:stepx}} - \mathbf{z}_{xy} \right] \\
    \gamma^2 p(x-1,y-1) & = \gamma p(x-1,y) + \gamma p(x,y-1) - p(x,y) + \mathbf{z}_{xy} \\
\end{split}
\label{eq:pxm1ym1steps}
\end{equation}

Moving $p(x,y)$ to the left and $\gamma^2 p(x-1,y-1)$ to the right, and we get:

\begin{equation}
    p(x,y) = \gamma p(x-1,y) + \gamma p(x,y-1) - \gamma^2 p(x-1,y-1) + \mathbf{z}_{xy}
\label{eq:parallel_recursion}
\end{equation}

Because we proved that $p(1, 1) = r(1, 1)$, and we proved that both $p(x,1) = r(x,1)$ and $p(1,y) = r(1,y)$, and because 
\begin{equation}
    r(x,y) = \gamma r(x-1,y) + \gamma r(x,y-1) - \gamma^2 r(x-1,y-1) + \mathbf{z}_{xy}
\end{equation}

With the parallel form in equation \ref{eq:parallel_recursion} being identical, we've proven that $p(x,y) = r(x,y)$ in the general case. QED.

\section{Relation to Retention in RetNet}\label{apdx:retnet_relation}

Recall that in RetNet \cite{sun2023retentive}, they define the recurrent retention formula as

\begin{equation}
\begin{aligned}
    s_n &= A s_{n-1} + K_n^\intercal v_n^{}, & A &\in \mathbb{R}^{d \times d}, & K_n &\in \mathbb{R}^{1 \times d} \\
    o_n &= Q_n s_n = \sum_{m=1}^{n} Q_n A^{n-m} K_m^\intercal v_m^{}, & & & Q_n &\in \mathbb{R}^{1 \times d}
\end{aligned}
\tag{RetNet Eq (1) \cite{sun2023retentive}}
\end{equation}

with

\begin{equation}
\begin{aligned}
    A &= \Lambda \left( \gamma e^{i \theta} \right) \Lambda^{-1} \\
    A^{n-m} &= \Lambda \left( \gamma e^{i \theta} \right)^{n-m} \Lambda^{-1} \\
            &= \Lambda \gamma^{n-m} e^{i(n-m)\theta} \Lambda^{-1}
\end{aligned}
\end{equation}

We note that $A$ combines xPos \cite{sun2022length} ($e^{i(n-m) \theta})$ with the $\gamma$ decay factor introduced in RetNet. If one wants to omit xPos, they can set $\theta = 0$, resulting in

\begin{equation}
\begin{aligned}
    A^{n-m} &= \Lambda \gamma^{n-m} e^{i(n-m)\cdot 0} \Lambda^{-1} \\
      &= \Lambda \gamma^{n-m} e^0 \Lambda^{-1} \\
      &= \Lambda \gamma^{n-m} I \Lambda^{-1} \\
      &= \gamma^{n-m}
\end{aligned}
\end{equation}

We then end up with $A=\gamma$, which is how we define it in \eqref{eq:rnn} for ViR, as we opt to use learned absolute positional embeddings instead. For future work, we plan on exploring xPos in two dimensions, recovering the relative position embedding via complex rotations.

\section{Impact of resolution scaling on throughput}

\begin{figure}[t]
    \centering
    \includegraphics[width=0.6\linewidth]{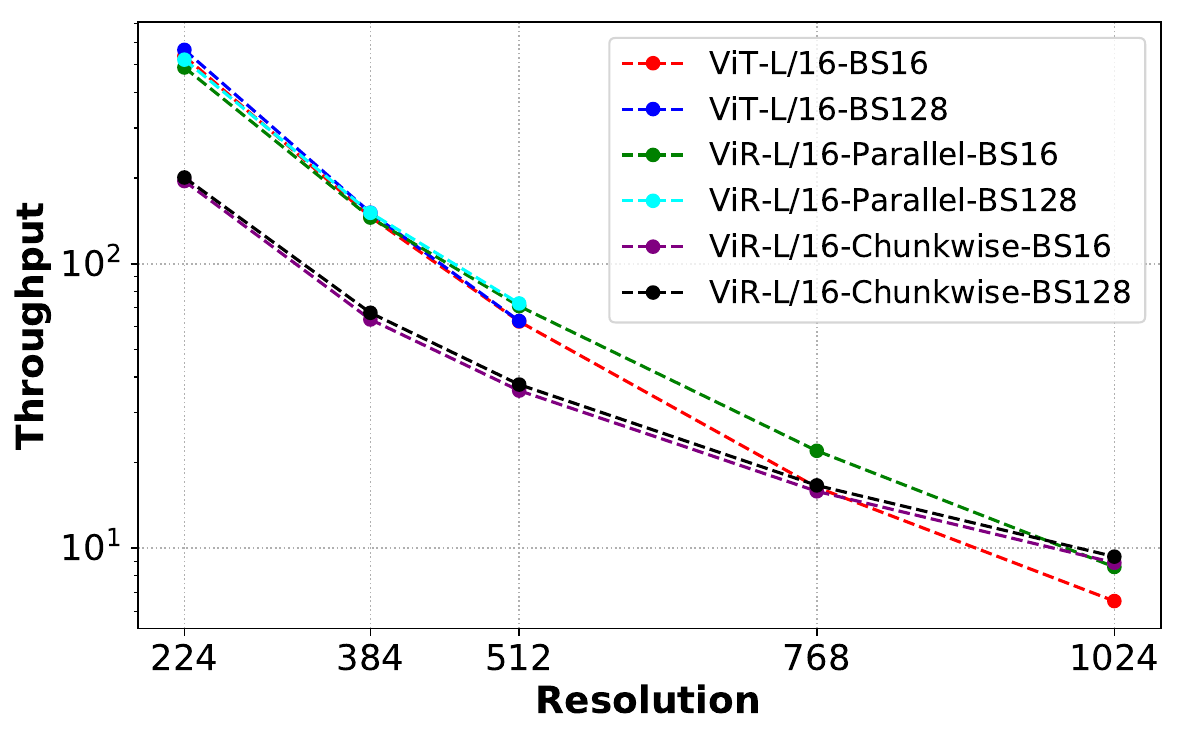}
    \caption{Effect of resolution scaling on image throughput for ViR-L/16 and ViT-L/16 models. Throughput is measured on an A100 80GB NVIDIA GPU with batch sizes of 16 and 128. For a batch size of 128, the memory is insufficient to process images for both ViT and parallel mode of ViR networks. For a batch size of 128 and $1024 \times 1024$ image resolution, ViR-L/16 with chunkwise formulation is the only configuration that can process images.}
    \label{fig:throughput_l}
\end{figure}

In Fig.~\ref{fig:throughput_l}, we investigate the impact of scaling the image resolution on throughput and demonstrate that ViR-L scales favorably compared to ViT-L counterpart. For a batch size of 16 and smaller image resolutions (\textit{e.g.} $224\times224$, ViT-L has a higher throughput when compared to ViR-L with chunkwise formulation. However, chunkwise formulation becomes comparable or faster at large image resolutions such as $1024\times1024$. In addition, with a batch size of 128, ViT-L cannot process images at $768\times768$ and $1024\times1024$ due to insufficient memory. However, ViR-L with chunkwise formulation can be leveraged to efficiently run images at these resolutions. These results validate the effectiveness of the proposed ViR as an effecicient and scalable model for processing high resolution image resolutions with larger batch sizes. 

\section{Architecture Details}
\label{sec:arch}
We present detailed architecture configuration of various HViR models in Table~\ref{table:arch-spec-abl}.

\begin{table*}[t]
\small
\centering
\addtolength{\tabcolsep}{-2pt}
\resizebox{.9\linewidth}{!}{
\begin{tabular}{c|c|c|c|c|c}
 & \begin{tabular}[c]{@{}c@{}}Output Size \\ (Downs. Rate)\end{tabular} & HViR-1  &HViR-2 & HViR-3 &  HViR-4 \\
\hline
\hline
\multirow{3}{*}{Stem} & \multirow{3}{*}{\begin{tabular}[c]{@{}c@{}}112$\times$112\\ (2$\times$)\end{tabular}} & $\begin{bmatrix}\text{Conv-BN-ReLU}\\\text{C:32, S:2}\end{bmatrix}$ $\times$ 1  & $\begin{bmatrix}\text{Conv-BN-ReLU}\\\text{C:64, S:2}\end{bmatrix}$ $\times$ 1  &$\begin{bmatrix}\text{Conv-BN-ReLU}\\\text{C:64, S:2}\end{bmatrix}$ $\times$ 1   & $\begin{bmatrix}\text{Conv-BN-ReLU}\\\text{C:64, S:2}\end{bmatrix}$ $\times$ 1  \\
\cline{3-6}
& & $\begin{bmatrix}\text{Conv-BN-ReLU}\\\text{C:80}\end{bmatrix}$ $\times$ 1   & $\begin{bmatrix}\text{Conv-BN-ReLU}\\\text{C:96}\end{bmatrix}$ $\times$ 1    & $\begin{bmatrix}\text{Conv-BN-ReLU}\\\text{C:128}\end{bmatrix}$ $\times$ 1   & $\begin{bmatrix}\text{Conv-BN-ReLU}\\\text{C:196}\end{bmatrix}$ $\times$ 1   \\
\hline
\multirow{3}{*}{Stage 1} & \multirow{3}{*}{\begin{tabular}[c]{@{}c@{}}56$\times$56\\ (4$\times$)\end{tabular}} & LN-2D, Conv, C:160, S:2  & LN-2D, Conv, C:192, S:2  & LN-2D, Conv, C:256, S:2 & LN-2D, Conv, C:392, S:2  \\
\cline{3-6}
& & $\begin{bmatrix}\text{ResBlock}\\\text{C:160}\end{bmatrix}$ $\times$1,  & $\begin{bmatrix}\text{ResBlock}\\\text{C:192}\end{bmatrix}$ $\times$ 3,   & $\begin{bmatrix}\text{ResBlock}\\\text{C:256}\end{bmatrix}$ $\times$ 3,   & $\begin{bmatrix}\text{ResBlock}\\\text{C:392}\end{bmatrix}$ $\times$ 3,   \\
\hline
\multirow{3}{*}{Stage 2} & \multirow{3}{*}{\begin{tabular}[c]{@{}c@{}}28$\times$28\\ (8$\times$)\end{tabular}} & LN-2D, Conv, C:320, S:2  & LN-2D Conv, C:384, S:2  & LN-2D, Conv, C:512, S:2  & LN-2D, Conv, C:768, S:2  \\
\cline{3-6}
& & $\begin{bmatrix}\text{ResBlock}\\\text{C:320}\end{bmatrix}$ $\times$ 3,   & $\begin{bmatrix}\text{ResBlock}\\\text{C:384}\end{bmatrix}$ $\times$ 3,    & $\begin{bmatrix}\text{ResBlock}\\\text{C:512}\end{bmatrix}$ $\times$ 3,   & $\begin{bmatrix}\text{ResBlock}\\\text{C:768}\end{bmatrix}$ $\times$ 3,   \\
\hline
\multirow{3}{*}{Stage 3} & \multirow{3}{*}{\begin{tabular}[c]{@{}c@{}}14$\times$14\\ (16$\times$)\end{tabular}} & LN-2D, Conv, C:640, S:2 & LN-2D, Conv, C:768, S:2  & LN-2D, Conv, C:1024, S:2   & LN-2D, Conv, C:1568, S:2 \\
\cline{3-6}
& & $\begin{bmatrix}\text{RetentionBlock}\\\text{C:640, head:8}\end{bmatrix}$ $\times$ 8 ,  & $\begin{bmatrix}\text{RetentionBlock}\\\text{C:768, head:8}\end{bmatrix}$ $\times$ 8,    & $\begin{bmatrix}\text{RetentionBlock}\\\text{C:1024, head:8}\end{bmatrix}$ $\times$ 12,   & $\begin{bmatrix}\text{RetentionBlock}\\\text{C:1568, head:16}\end{bmatrix}$ $\times$ 12,   \\
\hline
\multirow{3}{*}{Stage 4} & \multirow{3}{*}{\begin{tabular}[c]{@{}c@{}}7$\times$7\\ (32$\times$)\end{tabular}} & LN-2D, Conv, C:1280, S:2  & LN-2D, Conv, C:1536, S:2 & LN-2D, Conv, C:2048, S:2  & LN-2D, Conv, C:3136, S:2  \\
\cline{3-6}
& & $\begin{bmatrix}\text{RetentionBlock}\\\text{C:1280, head:16}\end{bmatrix}$ $\times$ 5,   & $\begin{bmatrix}\text{RetentionBlock}\\\text{C:1536, head:16}\end{bmatrix}$ $\times$ 5,    & $\begin{bmatrix}\text{RetentionBlock}\\\text{C:2048, head:16}\end{bmatrix}$ $\times$ 5,   & $\begin{bmatrix}\text{RetentionBlock}\\\text{C:3136, head:32}\end{bmatrix}$ $\times$ 5,   \\
\end{tabular}
}
\normalsize
\caption{Architecture detail of HViR models. BN and LN-2D denote Batch Normalization and 2D Layer Normalization, respectively.}
\label{table:arch-spec-abl}
\end{table*}

\subsection{Training Details}
For image classification experiments,  we used the ImageNet-1K dataset~\cite{deng2009imagenet} which contains 1.2 million images for training and 50,000 images for validations. All HViR models employ the LAMB optimizer~\cite{lamb} and are trained for 300 epochs with an initial learning rate of 5e-3 and a total batch size of 4096 using 32 NVIDIA A100 GPUs and Exponential Moving Average (EMA). In addition, we use standard data augmentation strategies similar to previous efforts~\cite{liu2022convnet,liu2021swin}. For semantic segmentation experiments, all models are trained on ADE20K dataset~\cite{zhou2017scene} dataset with UperNet network~\cite{xiao2018unified} and using Adam-W~\cite{loshchilov2017decoupled} optimizer with a learning rate of 6e-5 and batch size of 16. For object detection experiments, all models are trained on MS COCO dataset~\cite{lin2014microsoft} with Cascade Mask-RCNN~\cite{he2017mask} detection head with a 3 $\times$ schedule and use Adam-W~\cite{loshchilov2017decoupled} optimizer and a learning rate of 1e-4 and batch size of 16. 
\section{A simplified formulation of 2D Retention}

Recall the 2D recurrent formulation of retention

\begin{equation}
\begin{aligned}
\begin{split}
    r(1,1) & = \mathbf{z}_{11} \\
    r(x,1) & = \gamma r(x-1,1) + \mathbf{z}_{x1} \\
    r(1,y) & = \gamma r(1,y-1) + \mathbf{z}_{1y} \\
    r(x,y) & = \gamma r(x-1,y) + \gamma r(x,y-1) - \gamma^2 r(x-1,y-1) + \mathbf{z}_{xy}
\end{split}
\end{aligned}
\tag{\ref{eq:2d_retention_recurrent} revisited}
\end{equation}

It turns out that this is equivalent to the following formulation:

\begin{equation}
\begin{split}
    s_x(1,y) & = \mathbf{z}_{1y} \\
    s_x(x,y) & = \gamma s_x(x-1,y) + \mathbf{z}_{xy} \\
    \\
    s(x,1) & = s_x(x, 1) \\
    s(x,y) & = \gamma s(x,y-1) + s_x(x, y)
\end{split}
\label{eq:2d_ret_simple}
\end{equation}

\subsection{Proof}

First, we check that the base cases still hold between \eqref{eq:2d_retention_recurrent} and \eqref{eq:2d_ret_simple}.

\begin{equation}
    s(1,1) = s_x(1, 1) = \mathbf{z}_{11} = r(1, 1)
\end{equation}

\begin{equation}
    s(x,1) = s_x(x,1) = \gamma s_x(x - 1, y) + \mathbf{z}_{x1} = r(x,1)
\end{equation}

\begin{equation}
    s(1,y) = \gamma s(1, y-1) + s_x(1, y) = \gamma s(1,y-1) + \mathbf{z}_{1y} = r(1,y)
\end{equation}

The general case naturally arises from \eqref{eq:stepy}, which has that the difference between successive cells on the column is equal to the decayed sum of the current row. $s_x(x, y)$ is the recursive formulation of the sum in \eqref{eq:stepy}, and the proof of this follows from generalizing \eqref{eq:proof_first_row_parallel} to any row treated independently from previous rows.

\end{document}